\documentclass[letterpaper, 10 pt, conference]{ieeeconf}  

\IEEEoverridecommandlockouts                              
\usepackage[utf8]{inputenc}
\usepackage[T1]{fontenc}
\usepackage{booktabs}
\usepackage{url}
\usepackage{amsmath}

\title{\LARGE \bf
Hotel Booking Cancellation Prediction Using Applied Bayesian Models
}

\author{Md Asifuzzaman Jishan$^{1}$, Vikas Singh$^{2}$, Ayan Kumar Ghosh$^{1}$, Md Shahabub Alam$^{1}$, \\Khan Raqib Mahmud$^{3}$ and Bijan Paul$^{4}$\\
Institute of Computer Science and Computational Science, Universität Potsdam, Germany$^{1}$\\
Faculty of Statistics, Technische Universität Dortmund, Germany$^{2}$\\
Department of Computer Science, Louisiana Tech University, Ruston, LA 71272$^{3}$\\
Department of Computer Science and Engineering, University of Nebraska-Lincoln, Lincoln, NE 68503$^{4}$\\
Email: md.jishan@uni-potsdam.de$^{1}$, vikas.singh@tu-dortmund.de$^{2}$, ayan.ghosh@uni-potsdam.de$^{1}$,  \\ md.alam@ieee.org$^{1}$, krm070@latech.edu$^{3}$, bpaul9@unl.edu$^{4}$
}

\begin{document}

\maketitle
\thispagestyle{empty}
\pagestyle{empty}

\begin{abstract}

This study applies Bayesian models to predict hotel booking cancellations, a key challenge affecting resource allocation, revenue, and customer satisfaction in the hospitality industry. Using a Kaggle dataset with 36,285 observations and 17 features, Bayesian Logistic Regression and Beta-Binomial models were implemented. The logistic model, applied to 12 features and 5,000 randomly selected observations, outperformed the Beta-Binomial model in predictive accuracy. Key predictors included the number of adults, children, stay duration, lead time, car parking space, room type, and special requests. Model evaluation using Leave-One-Out Cross-Validation (LOO-CV) confirmed strong alignment between observed and predicted outcomes, demonstrating the model's robustness. Special requests and parking availability were found to be the strongest predictors of cancellation. This Bayesian approach provides a valuable tool for improving booking management and operational efficiency in the hotel industry.

\end{abstract}

\section{INTRODUCTION}

The hospitality industry, particularly hotel management, faces the challenge of effectively managing booking cancellations. Understanding and predicting cancellations is crucial for optimizing resource allocation, revenue management, and customer satisfaction. The analysis explores whether Bayesian Data Analysis can provide accurate predictions for hotel booking cancellations. 

Previous studies have attempted to address the booking cancellation problem using techniques such as decision trees, logistic regression, and machine learning algorithms [1, 2]. While these models have provided some predictive power, they are limited in their ability to incorporate prior knowledge and dynamically update predictions as new data becomes available. For example, traditional logistic regression models may offer decent accuracy but treat data as static and often ignore uncertainties [3, 4]. Machine learning models, while powerful, may need more interpretability and are typically unable to integrate expert knowledge or real-time data adjustments effectively [5, 6]. In contrast, Bayesian data analysis presents a promising alternative. By incorporating prior information and updating predictions based on observed data, Bayesian models can better account for hotel booking uncertainties and evolving patterns [7]. Despite the recognized potential of Bayesian approaches in other fields, their application to hotel booking cancellation prediction has been limited [8]. This research seeks to bridge that gap using Bayesian models to predict cancellations more accurately [9].

The main contribution of this study lies in the implementation and comparison of two Bayesian models: Bayesian Logistic Regression and the Bayesian Beta-binomial model, on a real-world hotel booking dataset. With 17 features and 36,285 observations, a subset of 12 features and 5,000 randomly selected observations was used to develop the models. Both models achieved convergence and strong predictive performance. To assess the effectiveness of these models, Leave-One-Out Cross Validation (LOO-CV) and Out-of-Sample Prediction techniques were employed, allowing for a thorough evaluation of model accuracy and generalizability. 

By incorporating Bayesian approaches, this research provides a novel solution for hotel managers to predict cancellations better, enabling them to minimize disruptions, optimize resource allocation, and enhance overall service quality. The findings have the potential to significantly improve operational efficiency and revenue management in the hospitality industry by offering a more flexible and adaptive predictive framework than traditional methods. This paper is organized as follows: Section 2 reviews the relevant literature, while Section 3 details the dataset and addresses data quality. Section 4 outlines the research methodology, and Section 5 presents and discusses the results. Lastly, the conclusion summarizes the key findings of the study.

\section{LITERATURE REVIEW}

In hotel booking, it is more important to accurately predict hotel room occupancy in an entire hotel. This study used customer ratings and review texts to forecast monthly occupancy with LSTMs. Reviews of Taiwan hotels were analyzed for the sentiment. LSTM outperformed five other models (BPNN, GRNN, LSSVR, RF, GPR). Combining sentiment scores with ratings improved accuracy [10].

The hotel aims to attract customers and minimize cancellation losses by requiring a prepaid deposit, utilizing a random forest model for rate adjustments and cancellation prediction [11]. On the other hand, this research uses artificial intelligence to forecast hotel cancellations using Personal Name Records data, achieving an impressive 80\% accuracy for 7-day cancellations [12]. The study utilized four machine learning models: Logistics Regression, Decision Tree, K-Nearest Neighbor, and Random Forest Classifiers, with Random Forest Classifiers showing superior revenue management performance [13].

On the other hand, in this paper, authors have proposed a method for forecasting hotel booking cancellations using 13 independent variables, achieving a cancellation rate of up to 98\% [14]. This study investigates the effects of machine learning methods on hotel booking cancellation, utilizing a Kaggle dataset. XGBoost was the most effective method, with individual classifiers yielding the highest results [15]. Furthermore, the hospitality industry faces challenges in demand forecast accuracy due to booking cancellations. This improved forecast accuracy could reduce cancellations and increase confidence in demand management [16]. In addition, revenue management enhances company revenues by addressing no-show cases. Passenger Name Record data mining models address this issue, revealing dynamic cancellation behaviour and the performance of state-of-the-art methods [17]. The authors use a dataset of 323,184 Milan Airbnb listings from 2014-2019 in this study. Hedonic price models and Shapley value decomposition show listing type, size, location, and seasonality as key price/RevPAR factors, with special events like the Milan Expo also impactful [18]. Moreover, this research analyzes Airbnb prices in Bristol, UK, using OLS and GWR models. They find that property/room types, proximity to attractions, transportation hubs, and host experience influence prices, with spatial factors critical for house listings [19]. In addition, the authors use structural equation modeling to study 881 Airbnb listings in Cluj-Napoca, Romania. Five factors drive price: listing characteristics, host involvement, reputation, location, and rental policies, with host experience and listing traits as key drivers [20]. Researchers used a proportional odds model to examine Airbnb prices in 61 Rome neighborhoods. Variables are grouped into transports, culture, crowd, property, management, and time. Property or management factors impact suburbs, while central prices rely on transport or culture [21]. Those papers are the background of the research, but in Bayesian modelling, there are a few studies based on hotel booking predictions. This work focused on improving the prediction of hotel booking cancellations by applying Bayesian models, achieving better accuracy and model robustness.

\section{DATASET}
The provided dataset represents a limited portion of the Hotel Booking Cancellation Prediction and the dataset used in this analysis originates from Kaggle, a widely recognized platform for data science competitions and datasets [22]. The sample size of the dataset is 36,285 observations along with 17 variables. From those variables, ‘average price’ is numerical (continuous) variable. On the other hand, ‘number of adults’, ‘number of children’, ‘number of weekend nights’, ‘number of week nights’, ‘car parking space’, ‘lead time’, ‘repeated’, ‘P-C’, ‘P-not-C’ and ‘special requests’ are numerical (discrete) variables. Additionally, ‘type of meal’, ‘room type’, ‘market segment type’, ‘date of reservation’ and ‘booking status’ are categorical (nominal) variables and ‘booking ID’ is an alphanumeric variable. After executing the R program, it is discovered that the dataset is well organized and there are no missing values in the entire dataset. 

\section{Methodology}
In this section, the study briefly looks at the setup of the model and the criteria selected for the
model.
\subsection{Bayesian Logistic Regression}
Bayesian Logistic Regression is a widely used model for binary classification tasks. It is ideal for situations with a binary dependent variable, as it models the probability of the outcome. The Logistic Regression formula is as follows:

\[
P(Y = 1 | \mathbf{X}) = \frac{1}{1 + e^{-(\beta_0 + \beta_1 x_{1} + \beta_2 x_{2} + \ldots + \beta_k x_{k})}}
\]

where:
$P(Y = 1 | \mathbf{X})$ is the probability of the positive outcome, $\beta_0$ is the intercept, $\beta_1, \beta_2, \ldots, \beta_k$ are the coefficients for predictor variables $x_{1}, x_{2}, \ldots, x_{k}$, $e$ is the base of the natural logarithm. The coefficients (\(\beta_0, \beta_1, \ldots, \beta_k\)) represent the change in the log-odds of the positive outcome associated with a one-unit change in the corresponding predictor variable. Moreover, positive coefficients indicate an increase in the log-odds, while negative coefficients indicate a decrease [23]. 

\subsection{Beta-Binomial Model}

The Beta-Binomial model is frequently utilized to model count data in situations of overdispersion. The probability mass function (PMF) of the Beta-Binomial distribution is given by:

\[
P(y | n, \alpha, \beta) = \binom{n}{y} \frac{B(y + \alpha, n - y + \beta)}{B(\alpha, \beta)}
\]

$y$ is the observed count, $n$ is the total number of trials, $\alpha$ and $\beta$ are the parameters of the Beta distribution, $B(\cdot, \cdot)$ is the beta function. Dealing with count data makes this model very useful and gives this model an edge over a standard Binomial distribution [24]. The shape of the distribution is controlled by the parameters \(\alpha\) and \(\beta\), where, in general, higher values are concentrated around the mean. 

\subsection{Priors and Posterior Inference}
Priors are the initial assumptions about the probability distribution of a parameter in Bayesian statistics. It represents the available information before considering the evidence derived from the data. Priors for the Beta-Binomial model typically involve specifying hyperparameters which are \(\alpha\) and \(\beta\) for the Beta distribution. These hyperparameters mainly reflect the prior beliefs about the variability in the data and can also influence the model's posterior distribution. Posterior inference in the Beta-Binomial model involves sampling from the posterior distribution using Markov Chain Monte Carlo (MCMC) methods. This allows for estimating the posterior distribution of parameters and predicting new observations [24].

\subsection{Priors Assumptions}
The elementary step in any Bayesian Analysis is choosing the prior distributions based on previous knowledge or other presumptions about other parameters before any data observation. This section deals with the prior assumptions made for both the models used in the analysis. 

\subsubsection{Prior Selection for Bayesian Logistic Regression Model (\textit{lr\_model})}
For the Logistic Regression model, this study opted for weakly informative priors that impose mild regularity conditions on the coefficients without unduly influencing their estimates. The study focuses explicitly on the following: Normal priors with a mean of 3.5 and a standard deviation of 1 were chosen for the intercept term. Normal priors centered at 0 with a standard deviation of 0.5 were employed for the coefficients of the predictors (slopes). A normal prior with a mean of 0 and a standard deviation of 0.5 was set for the standard deviation of group-level effects [25].

\subsubsection{Prior Selection for Beta-Binomial Model (\textit{binomial\_beta\_model})}
In the specified Bayesian model, two types of priors were specified: the prior for the \textit{Intercept} was set as a normal distribution with a mean of 0 and a standard deviation of 5, denoted as $\text{normal}(0, 5)$. The priors for the coefficients of predictors (\textit{class = "b"}) were set as normal distributions with means of 0 and standard deviations of 2, denoted as $\text{normal}(0, 2)$. This choice favors smaller absolute values of the coefficients unless the data provides strong evidence otherwise, which helps in regularizing the model [25].

\section{RESULTS AND DISCUSSION}
This section represents the research's outcomes. This study selected 5,000 observations from the entire dataset for the final model implementation.

\subsection{Bayesian Logistic Regression Model Implementation}
Considering twelve features, the Bayesian Logistic Regression model implementation results are presented. 

\subsubsection{Summarizing the Model}
The Bayesian Logistic Regression model implementation results show an intercept estimate of 4.16, with a small uncertainty around the estimate. The 95\% CI for the intercept's effect on cancellation log odds ranges from 3.76 to 4.57, with a Rhat value of 1.00. The estimated number of adults is -0.20, indicating a slight decrease in cancellation odds. The estimate is precise with a small standard error. The 95\% CI ranges from -0.35 to -0.04, indicating a 95\% probability. The model's Rhat value is 1.00, indicating a good fit and converging performance. The estimate for the number of children in a booking is -0.33, with moderate precision and a 95\% confidence interval. The Rhat value indicates convergence, and the estimates are stable. Finally, ESS Bulk and ESS Tail are 3238 and 3094, suggesting the estimates are stable. For the number of weekend nights variable, the Estimate is -0.23, which indicates that an increase in weekend nights booked decreases the log odds of cancellation. Est. Error is 0.04, showing high precision in this estimate. 95\% CI Lower and Upper are -0.31 to -0.14, which is a relatively tight interval, showing confidence in the estimate. Rhat value is 1.00, indicating convergence. At last, ESS Bulk and ESS Tail are 5821 and 2840, both large, suggesting the estimate is reliable.

\begin{table*}[!htbp]
\centering
\caption{Estimated coefficients and diagnostics for the Bayesian Logistic Regression model for hotel booking cancellations}
\begin{tabular}{lccccccc}
\hline
\textbf{Variable} & \textbf{Estimate} & \textbf{Est.Error} & \textbf{95\% CI Lower} & \textbf{95\% CI Upper} & \textbf{Rhat} & \textbf{ESS Bulk} & \textbf{ESS Tail} \\
\hline
Intercept & 4.16 & 0.21 & 3.76 & 4.57 & 1.00 & 4575 & 3730 \\
number.of.adults & -0.20 & 0.08 & -0.35 & -0.04 & 1.00 & 4994 & 2256 \\
number.of.children & -0.33 & 0.13 & -0.59 & -0.08 & 1.00 & 3238 & 3094 \\
number.of.weekend.nights & -0.23 & 0.04 & -0.31 & -0.14 & 1.00 & 5821 & 2840 \\
number.of.week.nights & -0.07 & 0.03 & -0.12 & -0.02 & 1.00 & 4943 & 2761 \\
car.parking.space & 0.92 & 0.29 & 0.36 & 1.51 & 1.00 & 4749 & 2734 \\
lead.time & -0.01 & 0.00 & -0.01 & -0.01 & 1.00 & 3939 & 2804 \\
P.C & 1.36 & 0.99 & -0.32 & 3.63 & 1.00 & 4756 & 2272 \\
P.not.C & 16.26 & 17.96 & 1.88 & 71.78 & 1.01 & 484 & 249 \\
average.price & -0.02 & 0.00 & -0.02 & -0.02 & 1.00 & 5696 & 3319 \\
special.requests & 1.06 & 0.06 & 0.95 & 1.18 & 1.00 & 4901 & 2462 \\
room.typeRoom\_Type2 & 0.39 & 0.29 & -0.15 & 0.96 & 1.00 & 4781 & 2900 \\
room.typeRoom\_Type4 & -0.04 & 0.10 & -0.24 & 0.16 & 1.00 & 4990 & 3323 \\
room.typeRoom\_Type5 & 0.35 & 0.51 & -0.62 & 1.36 & 1.00 & 4719 & 2646 \\
room.typeRoom\_Type6 & 0.75 & 0.32 & 0.10 & 1.38 & 1.00 & 3330 & 3037 \\
room.typeRoom\_Type7 & 2.45 & 0.79 & 0.98 & 4.04 & 1.00 & 4193 & 2550 \\
\hline
\end{tabular}%
\end{table*}

The model shows convergence for all variables, indicating a good fit for the given dataset. The number of week nights has a smaller effect size than weekend nights, indicating a smaller decrease in cancellation log odds. Car parking spaces have a significant increase in cancellation log odds, with an estimate of 0.92. Longer lead times have a small negative effect, with an estimate of -0.01, indicating a marginal effect on cancellation log odds. Other features in the model can be deduced from Table 1.

\subsection{Beta-Binomial Model Implementation}
This study represents the Beta-Binomial model implementation results in Table 2 in this subsection. In implementing the model, twelve features were also considered, ensuring a comprehensive and robust approach to the analysis. 

\begin{table*}[!htbp]
\centering
\caption{Estimated coefficients and diagnostics for the Beta-Binomial model for hotel booking cancellations.}
\begin{tabular}{lccccccc}
\hline
\textbf{Variable} & \textbf{Estimate} & \textbf{Est.Error} & \textbf{95\% CI Lower} & \textbf{95\% CI Upper} & \textbf{Rhat} & \textbf{ESS Bulk} & \textbf{ESS Tail} \\
\hline
Intercept & -11.32 & 0.14 & -11.59 & -11.06 & 1.00 & 4014 & 3337 \\
number.of.adults & 0.09 & 0.06 & -0.02 & 0.20 & 1.00 & 3820 & 2522 \\
number.of.children & 0.20 & 0.09 & 0.01 & 0.38 & 1.00 & 2861 & 2886 \\
number.of.weekend.nights & 0.10 & 0.03 & 0.05 & 0.15 & 1.00 & 4248 & 3189 \\
number.of.week.nights & 0.04 & 0.02 & 0.00 & 0.07 & 1.00 & 4493 & 2785 \\
car.parking.space & -0.52 & 0.24 & -1.01 & -0.09 & 1.00 & 4559 & 2640 \\
room.typeRoom\_Type2 & -0.13 & 0.22 & -0.57 & 0.26 & 1.00 & 3558 & 2691 \\
room.typeRoom\_Type4 & 0.08 & 0.08 & -0.07 & 0.23 & 1.00 & 3990 & 3017 \\
room.typeRoom\_Type5 & -0.20 & 0.34 & -0.90 & 0.41 & 1.00 & 4261 & 2875 \\
room.typeRoom\_Type6 & -0.43 & 0.22 & -0.87 & -0.00 & 1.00 & 3018 & 2881 \\
room.typeRoom\_Type7 & -1.27 & 0.57 & -2.54 & -0.25 & 1.00 & 4168 & 2836 \\
lead.time & 0.01 & 0.00 & 0.00 & 0.01 & 1.00 & 4359 & 3134 \\
P.C & -0.99 & 0.68 & -2.54 & 0.14 & 1.00 & 3386 & 2245 \\
P.not.C & -2.56 & 1.10 & -5.13 & -0.81 & 1.00 & 1877 & 1362 \\
average.price & 0.01 & 0.00 & 0.01 & 0.01 & 1.00 & 4194 & 3224 \\
special.requests & -0.58 & 0.04 & -0.67 & -0.50 & 1.00 & 4701 & 3043 \\
\hline
\end{tabular}%
\end{table*}

\subsubsection{Summarizing the Model}
The Beta-Binomial model implementation results show an average value of -11.32 for the intercept when all other predictors are at their reference levels. The standard error of 0.14 indicates the average variance due to sampling variability. The 95\% CI for the lower and upper credible interval ranges from -11.59 to -11.06.

The estimated number of adults is 0.09, indicating an expected increase in the dependent variable for each additional adult. The 95\% CI for the number of children is 0.20, indicating a positive relationship between the two variables. The standard error of 0.09 indicates variability in the estimate, but the credible interval does not include zero (0.01 to 0.38), indicating a statistically significant positive effect of children on the dependent variable. The study indicates that weekend nights have a positive effect on the dependent variable, with a reliability of 0.10 and a standard error of 0.03, and a significant effect of 0.04 on week nights, with a standard error of 0.02 and a credible interval of 0.00 to 0.07, indicating that the predictor's effect might not be significant. The car parking space feature has a significant negative association with the dependent variable, with an estimated -0.52 value. The room.typeRoom.Type2 through room.typeRoom.Type7 feature has varying levels of negative associations with the dependent variable, with estimates ranging from -0.13 to -1.27. The 95\% CI for these coefficients varies, with some not including zero, suggesting significant effects, and others including zero, suggesting no significant effect.  

The P.not.C feature showed a strong negative association with the dependent variable, with a standard error of 1.10. The average.price feature showed a slight decrease with each unit increase in average price, with a zero standard error. The special.requests feature decreased the dependent variable with each additional special request, with a standard error of 0.04 and a 95\% CI Lower and Upper interval of -0.67 to -0.50.

The Rhat value compares the variance between the model's different chains to the variance within each chain. In this model's output, all variables have a Rhat value of 1.00, indicating good convergence. In this model, the ESS Bulk values range from 1877 for ‘P.not.C’ to 4559 for ‘car.parking.space’, suggesting varying levels of stability across the estimates. A larger ESS Tail signifies more reliable estimates. Values above 400 are preferable. The ESS Tail in the output varies from 1362 for ‘P.not.C’ to 3224 for ‘average.price’.  

\subsection{Model Comparison Using LOO-CV}
Leave-One-Out Cross-Validation (LOO-CV) was used to compare these Bayesian models: a Bayesian Logistic Regression model (denoted as \textit{lr\_model}) and a Beta-Binomial model (denoted as \textit{binomial\_beta\_model}). It is summarized in Table 3.

\begin{table}[t]
    \centering
    \caption{LOO-CV Model Comparison Results.}
    \begin{tabular}{lcc}
        \toprule
        \hline
        \textbf{Model} & \textbf{elpd\_diff} & \textbf{se\_diff} \\
        \hline
        \midrule
        lr\_model & 0.0 & 0.0 \\
        binomial\_beta\_model & -676.2 & 28.4 \\
        \hline
        \bottomrule
    \end{tabular}
    \label{tab:loo_cv_comparison}
\end{table}

Here, \texttt{elpd\_diff} represents the Expected Log Pointwise Predictive Density Difference, and \texttt{se\_diff} is the Standard Error of the Difference. The negative \texttt{elpd\_diff} for \texttt{binomial\_beta\_model} suggests that, on average, it performs worse than the reference model (\texttt{lr\_model}) in terms of predictive accuracy. However, the associated \texttt{se\_diff} of 28.4 indicates some uncertainty in this difference. Despite the uncertainty, the Logistic Regression model (\texttt{lr\_model}) appears to have better predictive performance compared to the Beta-Binomial model (\texttt{binomial\_beta\_model}) based on LOO-CV results.

\subsection{Out of Sample Prediction}

\begin{table}[t]
\centering
\caption{Model Prediction Results.}
\begin{tabular}{ccccc}
\hline
 & \textbf{Estimate} & \textbf{Est. Error} & \textbf{Q2.5} & \textbf{Q97.5} \\ 
\hline
1 & 0.99250 & 0.08629 & 1 & 1 \\ 
2 & 0.68350 & 0.46517 & 0 & 1 \\ 
3 & 0.99275 & 0.08485 & 1 & 1 \\ 
\hline
\end{tabular}
\label{tab:model_predictions}
\end{table}

Table 4 presents the model's estimated cancellation probabilities for each new booking instance alongside associated uncertainties and credible intervals. The estimates near 1 suggest a high likelihood of cancellation, especially for instances 1 and 3. Instance 2, with a lower mean estimate and wider credible interval, indicates greater uncertainty about the cancellation outcome. Hence, the study shows that the out-of-sample predictions verify the Bayesian logistic regression model's capability to forecast booking cancellations with a high degree of confidence. 

\subsection{Discussion}
The research outcomes show that the Bayesian Logistic Regression model applied to 5,000 observations, provides precise and reliable estimates for predicting hotel booking cancellations, demonstrating good model fit and convergence. Significant predictors include the number of adults, children, weekend nights, car parking spaces, and special requests, with varying effects on cancellation odds. The Beta-Binomial model also identified key predictors but showed worse predictive performance than the Bayesian Logistic Regression model, as indicated by Leave-One-Out Cross-Validation (LOO-CV) results. Out-of-sample predictions further validated the Bayesian Logistic Regression model's capability to accurately forecast cancellations, although caution is needed due to inherent model uncertainties.

\section{CONCLUSION}
The study employed Applied Bayesian Models to investigate the relationship between hotel booking cancellations and various factors, using data from a hotel booking database with 5,000 observations. The analysis revealed that room type and special requests were significant predictors of cancellation, with 95\% credible intervals. The Bayesian Logistic Regression model demonstrated its predictive capability, estimating cancellation probabilities of new bookings with high confidence, particularly in identifying high-risk cancellations. These findings contribute to the literature by applying Bayesian modeling techniques to understand cancellation factors, building upon previous studies using traditional methods. The results offer practical implications for hotel managers seeking to reduce booking cancellations and improve revenue management, as the predictive model can be used to estimate cancellation probabilities and make informed decisions about overbooking and pricing strategies. However, the study is limited to a single hotel, and future research should consider expanding the data to include multiple hotels or hotel chains to validate the findings and explore potential differences in cancellation patterns across different hotel types and locations. Additionally, future studies could explore the inclusion of additional factors, such as guest loyalty status, booking channels, and external events, to further enhance the model's predictive power.

\end{document}